# 🔍 See Different, Think Better: Visual Variations Mitigating Hallucinations in LVLMs

Ziyun Dai
dzy223@shu.edu.cn
Shanghai University
Shanghai, China

Xiaoqiang Li*
xqli@shu.edu.cn
Shanghai University
Shanghai, China

Shaohua Zhang[†]
shaohuazhang@fudan.edu.cn
Shanghai Business School
Shanghai, China

Yuanchen Wu
yuanchenwu@shu.edu.cn
Shanghai University
Shanghai, China

Jide Li
iavtvai@shu.edu.cn
Shanghai University
Shanghai, China

## Abstract

Large Vision-Language Models (LVLMs) have demonstrated remarkable capabilities in visual understanding and multimodal reasoning. However, LVLMs frequently exhibit hallucination phenomena, manifesting as the generated textual responses that demonstrate inconsistencies with the provided visual content. Existing hallucination mitigation methods are predominantly text-centric, the challenges of visual-semantic alignment significantly limit their effectiveness, especially when confronted with fine-grained visual understanding scenarios. To this end, this paper presents **ViHallu**, a **Vi**sion-Centric **Hallu**cination mitigation framework that enhances visual-semantic alignment through **Visual Variation Image Generation** and **Visual Instruction Construction**. ViHallu introduces *visual variation images* with controllable visual alterations while maintaining the overall image structure. These images, combined with carefully constructed visual instructions, enable LVLMs to better understand fine-grained visual content through fine-tuning, allowing models to more precisely capture the correspondence between visual content and text, thereby enhancing visual-semantic alignment. Extensive experiments on multiple benchmarks show that ViHallu effectively enhances models' fine-grained visual understanding while significantly reducing hallucination tendencies. Furthermore, we release ViHallu-Instruction, a visual instruction dataset specifically designed for hallucination mitigation and visual-semantic alignment. Code is available at https://github.com/oliviadzy/ViHallu.

## 1 Introduction

Recently, Large Vision-Language Models (LVLMs) [14, 20, 21] have shown remarkable potential in tasks such as image description [14] and visual question answering [6, 21]. However, due to weak vision-text alignment, these models tend to generate hallucinations [16, 22, 45], manifesting as incorrectly identifying nonexistent objects, misattributing object properties, or misrepresenting spatial relationships. These issues not only diminish model reliability but also significantly constrain the practical application of LVLMs in critical domains such as medical diagnosis and autonomous driving.

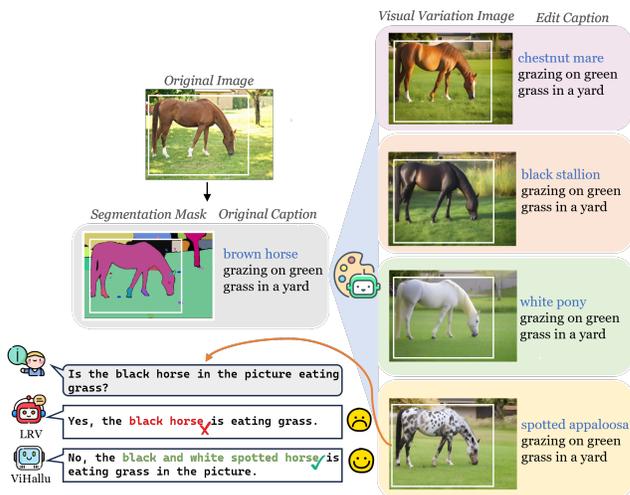

Figure 1: By inputting segmentation masks along with guiding caption into the controllable T2I model, the target visual variation image can be generated, maintaining structural similarity with the original image while exhibiting controlled local alterations.

The visual hallucination in LVLMs stems primarily from visual-semantic misalignment, where the model's textual outputs fail to accurately correspond to the visual elements present in the images. This cross-modal misalignment becomes particularly pronounced when models attempt to process fine-grained visual distinctions. For instance, models frequently struggle to differentiate between semantically similar scenes like *"a speckled black and white bird observing a chestnut brown horse eating grass in a meadow"* versus *"a chestnut brown bird eating grass while watching a speckled black and white horse standing in a meadow"*, despite the stark visual contrast these scenes represent. Such fundamental visual-semantic misalignments severely impair the visual understanding capability of LVLMs.

Previous research has primarily adopted text-centric approaches to mitigate hallucination in LVLMs, including improving the quality of the dataset through noise and error elimination [41] and the incorporation of negative textual examples [12, 18, 44]. LRV-Instruction [18] proposed enhancing the robustness of LVLMs against hallucinations by incorporating additional negative instructions. HalluciDoctor [41] focused on reducing hallucination impacts by detecting and eliminating hallucinated text content in visual instruction data. However, these text-centric strategies demonstrate limited efficacy in addressing fine-grained visual distinctions and do not sufficiently enhance the model's visual comprehension capabilities. As illustrated in Figure 8, when presented with a black-and-white spotted appaloosa horse, LRV misinterprets the prompt to indicate that the image contains a black horse. Consequently, developing vision-centric approaches to enhance visual-semantic alignment may be promising for addressing these visual hallucination.

---

*Corresponding author

To strengthen visual-semantic alignment in LVLMs, we introduce ***visual variation samples***, which exhibit fine-grained visual differences (such as scene elements, object categories, and attributes, etc. ), as shown in the right side of Figure 8. **By training models to recognize these fine-grained visual distinctions, we enable models to more precisely capture the correspondence between visual content and text, thereby enhancing visual-semantic alignment**. However, such visual variation samples rarely occur in natural datasets. This raises two critical questions: How can we effectively generate high-quality visual variation samples? And how can we leverage these samples to enhance the visual-semantic alignment of LVLMs?

Motivated by recent advances in Text-to-Image (T2I) generation [8, 15, 24, 27, 43], we present **ViHallu**, a **Vi**sion-Centric **Hallu**cination mitigation framework that enhances visual-semantic alignment through **Visual Variation Image Generation** and **Visual Instruction Construction**. At its core, ViHallu introduces a novel generation approach that combines text guidance with segmentation mask control to produce high-quality visual variation images that conform to specified captions and preserve the global structure of original images. The generated samples differ from their original counterparts only in regions corresponding to the modified textual caption. As shown in Figure 8, when replacing *"brown horse"* with *"chestnut mare"* in the image caption, only the target region transforms while maintaining structural consistency with the original image. **In the visual variation image generation process, we place objects in uncommon contextual settings where they rarely appear, creating counterfactual interventions [26]. Fine-tuning on these samples helps models rely on visual features rather than contextual expectations, reducing misleading correlations between frequently co-occurring objects. This improves the model's ability to make judgments based on visual evidence instead of statistical associations.** To leverage these samples effectively, we construct tailored visual instructions for paired original and variation images, enabling LVLMs to learn fine-grained discriminative features and enhance visual-semantic alignment through high-quality image-instruction pairs.

Our main contributions are summarized as follows:

- The **ViHallu** propose a novel visual variation image generation approach that integrates text guidance and segmentation mask control, producing samples that exhibit controlled visual alterations. while maintaining the overall image structure.

- To the best of our knowledge, **ViHallu** is the first to construct tailored instruction data for visual variation images, establishing a novel visual instruction data construction paradigm.

- We release **ViHallu-Instruction**, a visual instruction dataset specifically designed for fine-grained visual-semantic alignment. This dataset includes carefully curated visual variation images and high-quality instruction data to facilitate research in hallucination mitigation for LVLMs.

- Extensive empirical evaluations show that **ViHallu** significantly mitigates hallucination in LVLMs and enhances model robustness.

## 2 Related Work

### 2.1 Large Vision-Language Model

Large Language Models (LLMs) [2, 25, 33] have demonstrated remarkable performance across various Natural Language Processing (NLP) tasks. Similarly, LVLMs, which leverage the capabilities of LLMs, have undergone rapid development and shown significant advantages in multimodal tasks. The training of LVLMs typically comprises two critical phases: pre-training and visual instruction tuning. In the initial pre-training phase, the primary objective is to achieve vision-language alignment through large-scale image-text pair datasets. In the visual instruction tuning phase, the core objective is to enhance the model's ability to comprehend user instructions and complete specific tasks. Numerous visual instruction datasets are generated through the collaboration between LVLMs and LLMs. For instance, LLaVA [21] converts images into text descriptions with bounding box information and utilizes text-only GPT-4 [25] to generate new text data. Recently, with the emergence of the more powerful multimodal model GPT-4V [1], many works have adopted GPT-4V to generate data of higher quality, as exemplified by LVIS-Instruct4V [35] and ALLaVA [3].

### 2.2 LVLMs Hallucination

The hallucination problem in LVLMs has attracted much attention. Prior studies [12, 18, 36, 36, 41] have predominantly adopted text-centric approaches to mitigate hallucination in LVLMs. For instance, HACL [12] mitigates hallucination in LVLMs by introducing hallucinatory captions as hard negative text samples. HA-DPO [44] reframes hallucination mitigation as a preference optimization task by introducing hallucination-aware textual response pairs, training models to discriminate between hallucinated text and faithful text for the same visual input. However, these text-based approaches primarily focus on strengthening text generation capabilities while lacking enhancement of the model's visual understanding capabilities. This is particularly problematic when LVLMs need to distinguish subtle visual features. Several studies [11, 29, 31, 34] have demonstrated that incorporating model-generated negative images can enhance LVLMs multimodal compositional reasoning capabilities, suggesting the importance of the visual modality in improving model performance. Inspired by these advances and recognizing the limitations of purely text-centric approaches, we propose ViHallu, a flexible framework that strengthens visual-semantic alignment through the generation of visual variation image and visual instruction data. Unlike previous work that primarily focuses on textual domain, ViHallu emphasizes the visual domain by generating high-quality visual variation images.

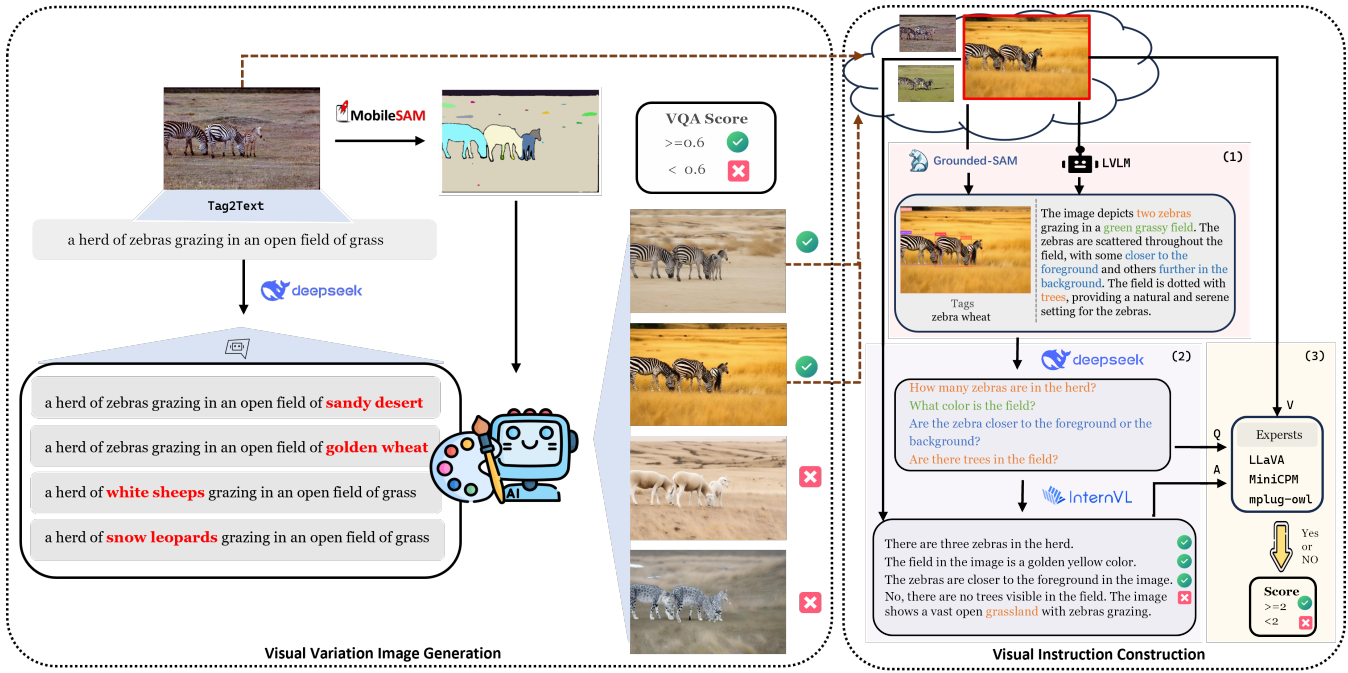

Figure 2: Overview of our framework ViHallu. The left shows the visual variation image generation process: (1) original image caption and segmentation mask generation, (2) caption editing through concept substitution, and (3) visual variation image generation with quality assessment. The right shows the instruction construction process: (1) detailed caption and object tag generation, (2) QA pair generation, and (3) QA pair quality assessment. Different types of hallucinations are indicated by distinct colors: Object Hallucination, Attribute Hallucination, and Relation Hallucination.

## 3 Methodology

This section delineates two phases of ViHallu. Visual variation image generation leverages advanced LLM and controllable T2I model to produce high-quality visual variation images. Visual instruction generation constructs instruction data based on the images.

### 3.1 Visual Variation Image Generation

This section introduces the visual variation image generation phase, which encompasses three primary modules: 1) original image caption generation and segmentation mask extraction, 2) caption editing through concept substitution, and 3) visual variation image generation and quality evaluation.

**Image Caption and Segmentation Mask.** To obtain image captions and segmentation masks from the original images, we employ Tag2Text [10] and MobileSAM [42] models. Tag2Text model extracts image tags and employs them to guide VLM in generating comprehensive image captions. MobileSAM demonstrates excellent zero-shot transfer capabilities, enabling precise object segmentation within images. By inputting the original image into MobileSAM, we obtain the segmentation mask image as shown in Figure 10, providing the foundation for visual variation image generation.

**Caption Editing.** To facilitate precise modifications in image captions for visual variation image generation, we propose a LLM-based caption editing mechanism. The editing mechanism generates variant captions that differ from the original captions exclusively in specified target object categories or attributes, while maintaining all other aspects of the original caption unchanged. The target object can encompass both individual object and background scene. To guide the model in generating new captions that meet these requirements, we design a specialized prompt template, as shown in Figure 11. In the implementation, considering comprehensive performance, we utilize the DeepSeek-chat V2 model [7] to execute the caption editing task. The primary purpose of this mechanism is to provide controlled inputs for generating visual variation images by precisely specifying which elements should be altered while preserving other contextual information. **Notably, this mechanism can integrate objects into tail scenes where they rarely appear, introducing counterfactual interventions**, as exemplified by the first caption on the left of Figure 10 "*a herd of zebras grazing in an open field of sandy desert*". Through such a caption, the T2I model produces corresponding images. During fine-tuning, this approach of combining rarely co-occurring scenes and objects helps reduce spurious correlations between frequently co-occurring objects [41].

> Replace one word in the given list with more detailed and visually relevant alternatives, and put the enriched discription to replace the object in the provided sentence. please notice that only replace one object.
> Return only the updated sentences, each separated by a new line.
>
> **Example:**
> **Words:**
> bird
> **Sentence:**
> A bird is sitting on a tree.
> **Answer:**
> A majestic eagle is sitting on a tree.
> A colorful parrot is sitting on a tree.
> A small sparrow is sitting on a tree.
> A large hawk is sitting on a tree.
>
> **Input:**
> **Words:**
> grass
> **Sentence:**
> A herd of zebras grazing in an open field of **grass**
>
> ---
>
> **Output:**
> A herd of zebras grazing in an open field of **green meadow**
> A herd of zebras grazing in an open field of **golden wheat**
> A herd of zebras grazing in an open field of **brown tundra**
> A herd of zebras grazing in an open field of **sandy desert**

Figure 3: Illustration of caption editing prompt design. The prompt gives a example of caption editing and takes a target word and an original caption as input, guiding the model to generate enriched variations by replacing the target object with more detailed alternatives. The example demonstrates how a basic reference "grass" is transformed into diverse, attribute-rich descriptions (such as "green meadow" and "golden wheat").

By pairing objects with unexpected scenes to create semantic inconsistencies, the model is forced to rely on visual evidence rather than statistical priors during the inference process.

**Visual variation Image Generation.** Through a text-guided and segmentation-mask-controlled generation model, target images are produced that maintain structural consistency while featuring controlled local variations. Utilizing previously obtained segmentation masks and captions as dual inputs: the masks ensure structural layout and object positioning in target images, while the target caption (differing only in specific object categories and attributes) guides content generation. We employ ControlNet++ [27] as the image generation model. Through the dual input of segmentation masks and target captions, this model accurately generates the desired images while maintaining natural visual coherence, with modified regions seamlessly integrated without artifacts such as edge discontinuities or composition errors. To ensure the quality of generated target images, we implement quality assessment using the VQAScore metric [17], which evaluates image-text alignment through the Visual Question Answering (VQA) model. A quality threshold is established by filtering out generated images with scores below 0.6, retaining only high-scoring images to maintain the generation quality standard. The filtered target images, together with the original images, constitute the image set.

### 3.2 Visual Instruction Generation

This section presents the high-quality visual instruction generation phase, which comprises three modules: 1) detailed description and object tag generation, 2) Question-Answer generation, and 3) quality assessment.

**Detailed Description and Object Tag Generation.** To construct fine-grained visual instruction data, we leverage LVLM's image-to-text capabilities to produce detailed image descriptions. Due to LVLM's inherent hallucination tendencies, the generated descriptions may contain hallucinations, as shown on the right of Figure 10, different types of hallucinations are indicated by distinct color. And the generated descriptions may not comprehensively cover all objects in the image, particularly in complex scenes where descriptions tend to focus on primary objects. To ensure that questions cover all objects, we employ Grounded-SAM [13, 23, 30] for object detection and segmentation, extracting object tags to guarantee comprehensive coverage of all objects in the image.

**Question-Answer Generation.** In the question generation phase, we leverage DeepSeek-chat V2 to process detailed descriptions and object tags to generate questions. Detailed descriptions provide contextual information and the presence of hallucinations in LVLM-generated descriptions is helpful for the detection and removal of hallucinations. **When the LLM processes a description containing hallucinations, it naturally exhibits a tendency to generate questions about these hallucinated elements**. As shown on the right of Figure 10, when a description incorrectly states there are trees in the image, this triggers questions such as *"Are there trees in the field?"*. Object tags ensure comprehensive coverage by identifying all objects in the images, including those absent from descriptions. The DeepSeek-chat V2 model generates questions based on these inputs, targeting areas where LVLMs are prone to hallucinate. To generate detailed and accurate answer for each question, we utilize InternVL2.5 [5] as the answer generation model, which responds to generated questions based on the actual image content rather than the potentially hallucinated description. Detailed prompt templates are provided in the Appendix.

**Quality Assessment.** To evaluate the quality of Question-Answer (QA) pairs, we employ a panel of LVLM experts (*e.g.*, LLaVA-1.5 [19], MiniCPM-V 2.6 [38], and mPLUG-OWL3 [39]). This multi-model consensus approach mitigates biases that might be present in individual models and enhances the robustness of our quality assessment process. Each QA pair and corresponding image are inputted to these expert models, which assess the alignment between the answers and the image content, producing binary outputs: *"Yes"* for aligned content and *"No"* for misaligned content. The experts models evaluate whether the answer accurately reflects what is visible in the image, checking for both factual correctness and comprehensiveness. We retain QA pairs endorsed by at least two expert models and discard those rejected by two or more experts. This majority voting mechanism ensures that only high-quality QA pairs are included in the final dataset. More details on the quality assessment methodology are provided in the Appendix.

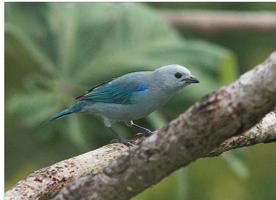
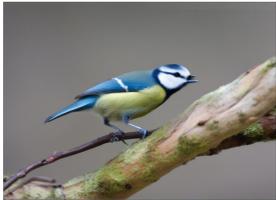
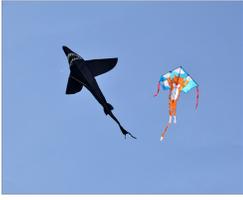
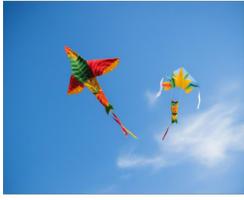

| Question | What color is the bird? | What color is the bird? | What is the shape of the smaller kite? | What is the shape of the smaller kite? |
|---|---|---|---|---|
| LLaVA-1.5 | The bird is blue and **white**. | The bird is blue and yellow. | The shape of the smaller kite is **a dog**. | The smaller kite is shaped like **a star**. |
| ViHallu-Instruction | The bird in the picture is primarily light blue with a subtle hint of gray. The most striking feature is the vibrant blue patch on its wings, which stands out against the softer blue and gray tones of its body. | The bird in the picture is predominantly blue and yellow. It has a bright blue cap and wings, with a white face and a yellow underside. The contrast between these colors makes it quite striking. | The smaller kite is designed to look like a cartoonish cat. It has an orange body with white and black stripes, and it appears to be in a playful pose with its arms and legs spread out. The kite has a blue background with white cloud patterns, adding to its whimsical appearance. | The smaller kite in the picture has a diamond shape with a long tail. The top part of the kite is shaped like a flower or sun with petals extending outward. The tail is elongated and has a design that matches the kite's colors. This combination gives it a dynamic and colorful appearance in the sky. |

Figure 4: Samples from the ViHallu-Instruction dataset. The top row displays two pairs of images: original images (left) and their corresponding visual variation images (right). Below are exemplar QA pairs from ViHallu-Instruction alongside responses from the baseline LLaVA-1.5 model. For clarity, incorrect responses are highlighted in red, while detailed responses are emphasized in blue.

## 4 Experiments

In this section, we present the ViHallu-Instruction dataset construction, baseline models, evaluation metrics, and implementation details.

### 4.1 Implementation Setup

The following describes the experimental setup, including dataset construction and baseline models.

**Dataset Construction.** The ViHallu-Instruction dataset consists of 6,770 images paired with approximately 50k (±10k) tailored instructions for the baseline models. Representative samples from ViHallu-Instruction are shown in Figure 9. Through the generation of visual variation images, 7,209 images are created and evaluated using LLaVA-1.5-13B for VQA scoring. This quality assessment yields 5,051 high-quality visual variation images, which are combined with their 1,719 corresponding original images to form the final dataset of 6,770 images. The visual instruction generation process relies on detailed image descriptions generated by the baseline model, specifically targeting hallucinations in model-generated descriptions to facilitate hallucination correction during fine-tuning. For question generation, the DeepSeek-chat V2 model extracts seven questions per image from the descriptions. To enhance the model's ability to distinguish between original and visual variation samples, a subset of questions generated from original samples is applied to visual variation images. As shown in Figure 9, the visual differences between the original and visual variation samples naturally lead to distinct answers for the same questions. During fine-tuning, these different responses help the model better identify distinctions between visual variation and original samples, thus mitigating fine-grained hallucinations. Answer generation utilizes InternVL-2.5-38B to produce responses based on image content. Quality assessment by three expert models filters out approximately 20% of the answers. More construction details and samples of ViHallu-Instruction are provided in the Appendix.

**Model Baselines.** For comprehensive evaluation, we employ three representative open-source LVLMs: (1) LLaVA 1.5 (7B) [19]; (2) MiniGPT4 v2 [4]; (3) Qwen2-VL (7B) [37].

### 4.2 Implementation Details

For fine-tuning, we construct training data using ViHallu, which maintains the same image content while incorporating model-specific visual instruction data tailored for each LVLM.

**LLaVA-1.5.** For LLaVA-1.5-7B, the full parameter fine-tuning is carried out with a batch size per device of 16. The implementation utilizes a cosine annealing learning rate scheduler with an initial learning rate of 2e-5.

**MiniGPT-4 v2.** The MiniGPT-4 v2 is fine-tuned with LoRA, with the rank set to 64 and $\alpha$ to 16, utilizing linear projection layers as the sole trainable modules. The optimization process employs a linear warmup with cosine annealing learning rate schedule at 1e-5.

**Qwen2-VL.** The Qwen2-VL model is fine-tuned using full-parameter tuning while keeping the vision tower frozen. Only the multimodal projection layers are trained, with a batch size of 4 and gradient accumulation steps of 1. The optimization process employs a cosine annealing learning rate schedule at 1e-5 with a 0.1 warmup ratio.

### 4.3 Evaluation Metric

The following presents the evaluation metrics employed in experiments.

**POPE**[40], the Polling-based Object Probing Evaluation, evaluates object hallucination in LVLMs about object presence in images through balanced Yes/No questions (i.e., 50% vs. 50%). The benchmark implements three progressively challenging negative sampling strategies: random selection from arbitrary objects, popular selection from high-frequency objects, and adversarial selection from scene-relevant co-occurring objects. Based on MSCOCO, A-OKVQA, and GQA datasets, POPE comprises 27,000 question-answer pairs, with performance evaluated using Accuracy, Precision, Recall, and F1 Score metrics.

Table 1: Results of LVLMs on *random*, *popular*, and *adversarial* settings of POPE dataset. The best results are shown in bold.

| Setting | Model | Accuracy (%) | F1 Score (%) |
|---|---|---|---|
| Random | LLaVA-1.5 | 87.90 | 87.35 |
| | w/ViHallu | **90.13** | **89.59** |
| | MiniGPT4 V2 | 82.30 | 84.07 |
| | w/ViHallu | **85.23** | **85.66** |
| | Qwen2-VL | 90.67 | 89.92 |
| | w/ViHallu | **91.07** | **90.44** |
| Popular | LLaVA-1.5 | 87.23 | 86.37 |
| | w/ViHallu | **88.40** | **87.98** |
| | MiniGPT4 V2 | 79.97 | 82.30 |
| | w/ViHallu | **85.20** | **85.59** |
| | Qwen2-VL | 89.33 | 88.64 |
| | w/ViHallu | **89.40** | **88.86** |
| Adversarial | LLaVA-1.5 | 84.43 | 84.28 |
| | w/ViHallu | **84.97** | **84.96** |
| | MiniGPT4 V2 | 73.07 | 77.61 |
| | w/ViHallu | **75.77** | **78.43** |
| | Qwen2-VL | **87.27** | **86.74** |
| | w/ViHallu | 86.93 | 86.61 |

Table 2: Results of hallucination mitigation methods on POPE dataset. We report the average accuracy and F1 score of POPE. The best results are shown in bold.

| Method | Accuracy (%) | F1 Score (%) |
|---|---|---|
| LLaVA-1.5 [19] | 86.52 | 86.00 |
| LLaVA-1.5$_{w/HA-DPO}$[44] | 86.63 | 86.87 |
| LLaVA-1.5$_{w/HACL}$[12] | 87.69 | 87.26 |
| LLaVA-1.5$_{w/VH}$[9] | / | 84.80 |
| LLaVA-1.5$_{w/RAR}$[28] | 87.16 | 86.43 |
| LLaVA-1.5$_{w/ViHallu(Ours)}$ | **87.83** | **87.51** |
| MiniGPT-4 v2 [4] | 78.45 | 81.33 |
| MiniGPT-4 v2$_{w/ViHallu(Ours)}$ | **82.07** | **83.23** |
| Qwen2-VL [37] | 89.09 | 88.43 |
| Qwen2-VL$_{w/ViHallu(Ours)}$ | **89.13** | **88.63** |

**LLaVA-Bench**[21] is an evaluation suite designed to assess LVLMs' capabilities across diverse visual tasks. The benchmark includes 24 images (including indoor and outdoor scenes, memes, paintings, sketches, etc.) with 60 questions spanning three categories: conversation (simple QA), detailed description, and complex reasoning. This benchmark evaluates models' ability to interpret various visual content in challenging tasks and generalizability to novel domains.

**MMHal-Bench**[32] is a specialized benchmark dataset designed to evaluate hallucination phenomena in MLLMs. The dataset comprises 96 image-question pairs, ranging in 8 question categories × 12 object topics. The eight question categories cover various types of hallucination, including object attributes, counting, spatial relations, etc. By leveraging the GPT-4o model for response analysis and scoring, MMHal-Bench provides a rigorous and practical evaluation framework for assessing the accuracy and reliability of LVLMs in real-world applications.

## 5 Results

This section presents a comprehensive evaluation of ViHallu through extensive experiments. We compare models fine-tuned with ViHallu against both their original baseline and other approaches for hallucination mitigation, demonstrating the effectiveness of our method across multiple benchmark tasks. Furthermore, additional analyses validate the reliability of our approach.

### 5.1 Evaluation on Benchmarks

**Results on POPE.** Experimental results on POPE under the *random*, *popular*, and *adversarial* settings are summarized in Table 1. Fine-tuning with ViHallu demonstrates significant reduction in hallucination for MiniGPT-4 v2, achieving accuracy improvements of 2.93%, 5.23%, and 2.70% on *random*, *popular*, and *adversarial* sets, respectively, accompanied by substantial improvements in F1 scores across all settings. For LLaVA-1.5, which exhibits relatively fewer hallucinations in its baseline version due to visual instruction fine-tuning during initial training, fine-tuning with ViHallu yields further performance improvements. The model achieves accuracy improvements of 2.23%, 1.17%, and 0.54% on *random*, *popular*, and *adversarial* sets respectively, along with enhanced F1 scores. Qwen2-VL exhibits the strongest performance among all baseline models, with initial accuracy scores of 90.67%, 89.33%, and 87.27% on *random*, *popular*, and *adversarial* settings respectively. After fine-tuning with ViHallu, Qwen2-VL shows further improvements in the *random* and *popular* settings, with accuracy increasing by 0.40% and 0.07% respectively. Corresponding F1 scores also improve by 0.52% and 0.22%. In the *adversarial* setting, Qwen2-VL shows a slight decrease in performance, with accuracy decreasing by 0.34% and F1 score by 0.13%. This small decrease can be attributed to minor bias effects that emerge during fine-tuning. Despite this decrease, Qwen2-VL's performance after fine-tuning still significantly outperforms both baseline and fine-tuned versions of the other models in all settings, demonstrating its exceptional robustness against object hallucinations.

The performance improvements across models on POPE after fine-tuning with ViHallu demonstrate enhanced detection of object hallucinations. Particularly noteworthy is the improvement in the *adversarial* setting for MiniGPT-4 v2 and LLaVA-1.5, where negative samples are drawn based on co-occurrence frequencies with true objects, which is traditionally the most challenging scenario for hallucination detection. The improvements in this setting validate our hypothesis that **ViHallu effectively mitigates hallucination by introducing counterfactual samples to reduce spurious correlations between frequently co-occurring objects**. The slight decrease for Qwen2-VL in this setting suggests that it already has a strong capability to identify counterfactual examples, making the introduction of such samples largely ineffective.

Table 3: Results of LVLMs on LLaVA-Bench (In-the-Wild) dataset. The best results are shown in bold.

| Model | Conversation | Detail description | Complex reasoning | All |
|---|---|---|---|---|
| LLaVA-1.5 | 58.3 | 49.6 | 69.5 | 59.3 |
| LLaVA-1.5$_{w/ViHallu}$ | **62.3** | **57.0** | **77.0** | **67.7** |
| MiniGPT4 V2 | 43.5 | 42.8 | 64.6 | 52.8 |
| MiniGPT4 V2$_{w/ViHallu}$ | **53.9** | **48.2** | **73.8** | **61.5** |
| Qwen2-VL | 68.5 | 69.0 | 70.6 | 69.6 |
| Qwen2-VL$_{w/ViHallu}$ | **87.5** | **76.0** | **86.3** | **84.0** |

Table 4: Results of LVLMs on MMHal-Bench dataset. The best results are shown in bold.

| Model | Overall Score↑ | Hallucination Rate↓ |
|---|---|---|
| LLaVA-1.5 | 1.93 | 0.70 |
| LLaVA-1.5$_{w/ViHallu}$ | **2.11** | **0.66** |
| MiniGPT4 V2 | 1.54 | 0.76 |
| MiniGPT4 V2$_{w/ViHallu}$ | **1.91** | **0.70** |
| Qwen2-VL | 2.94 | 0.51 |
| Qwen2-VL$_{w/ViHallu}$ | **3.64** | **0.48** |

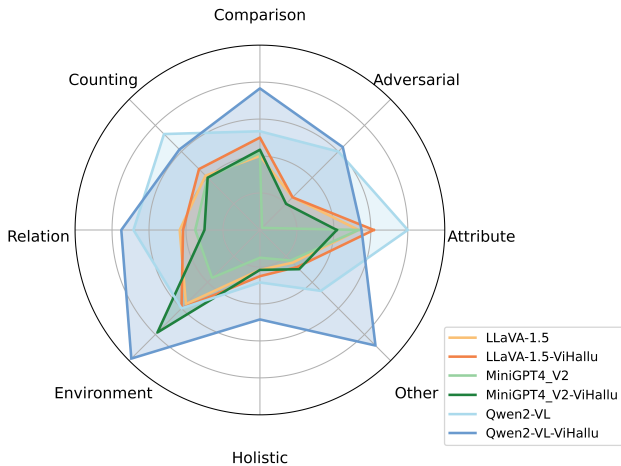

Figure 5: Performance comparison of different models before and after ViHallu fine-tuning across eight hallucination categories on MMHal-Bench.

Additionally, we compare ViHallu with other hallucination mitigation methods, with all approaches based on LLaVA-1.5-7B to ensure fair comparison, as shown in Table 2. The results show that ViHallu outperforms other methods, achieving the highest accuracy and F1 score on the POPE benchmark.

**Results on LLaVA-Bench.** To assess the impact of ViHallu on models' comprehensive capabilities, we conduct evaluations using LLaVA-Bench, a non-hallucination-focused benchmark. As shown in Table 3, the models demonstrate improved overall performance after fine-tuning with ViHallu. Specifically, MiniGPT-4 V2 exhibits performance improvements of 10.4, 5.4, and 9.2 points in *conversation*, *detailed description*, and *complex reasoning categories*, respectively. LLaVA-1.5 also shows enhancements of 4.0, 7.4 and 7.5 points. Most notably, Qwen2-VL shows the most substantial improvements with gains of 19.0, 7.0, and 15.7 points across the three evaluation categories, resulting in an impressive 14.4 point increase in overall performance. The results demonstrate that ViHallu effectively enhances models' conversational capabilities and fine-grained understanding through fine-tuning, achieved by incorporating visual variation samples and detailed instructions. These consistent improvements across all baseline models confirm that the ViHallu successfully strengthens visual-semantic alignment, enhancing models' ability to accurately interpret and reason about visual content.

**Results on MMHal-Bench.** Experimental results on MMHal-Bench are summarized in Table 4. For LLaVA-1.5, the application of ViHallu improves the overall score from 1.93 to 2.11, representing a 9.3% increase in performance. Simultaneously, its hallucination rate decreases from 0.70 to 0.66, achieving a 5.7% reduction in erroneous outputs. MiniGPT4 V2 demonstrates even more substantial improvements with ViHallu enhancement, with its overall score rising from 1.54 to 1.91, along with a significant reduction in hallucination rate from 0.76 to 0.70. Qwen2-VL shows the most impressive performance both before and after enhancement. Its baseline score of 2.94 already outperforms other models, and after applying ViHallu, this score increases to 3.64. Similarly, its already low hallucination rate further decreases from 0.51 to 0.48, representing a 5.9% reduction. The Figure 5 provides additional insights into model performance across eight hallucination categories. All models with ViHallu enhancement display expanded polygons in the radar chart, indicating improvements across most evaluation dimensions, while all enhanced models show notable improvements in previously challenging areas like *Adversarial* and *Environment* categories. These visualization results further confirm that ViHallu effectively enhances multimodal understanding and reduces various types of hallucinations across different model architectures.

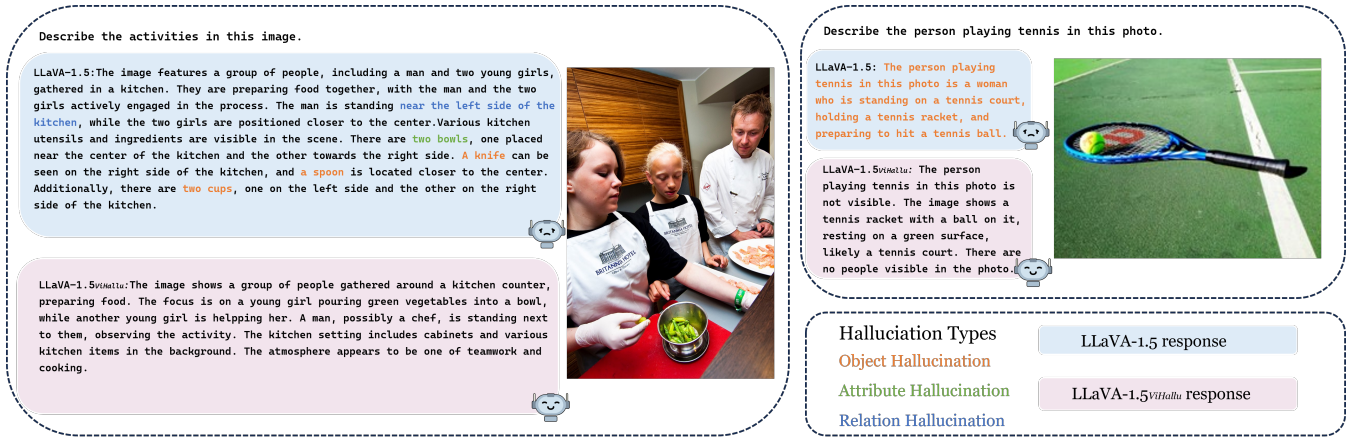

Figure 6: Response comparison between baseline LLaVA-1.5 and LLaVA-1.5$_{ViHallu}$ (LLaVA-1.5 fine-tuned with ViHallu) on two samples from MMHal-Bench dataset.

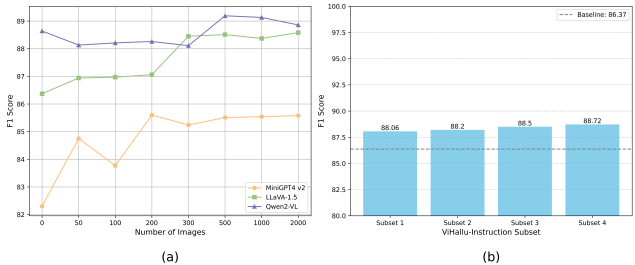

Figure 7: (a) F1 score on POPE benchmark of LVLMs fine-tuned with varying sizes of ViHallu-Instruction subsets. (b) F1 score on POPE benchmark of LLaVA-1.5 fine-tuned on four independent subsets of ViHallu-Instruction.

## 5.2 Further Analysis

**Case Study.** Figure 12 presents a comparison of responses between the baseline LLaVA-1.5 and LLaVA-1.5$_{ViHallu}$ fine-tuned with ViHallu. In the left example, the LLaVA-1.5 hallucinates multiple non-existent objects (such as cups, spoons, and knives) and makes spatial relationship errors (describing the man on the right as being on the left). In contrast, LLaVA-1.5$_{ViHallu}$ produces concise, hallucination-free response while effectively preventing interference from frequently co-occurring objects. In the right example, when asked about a non-existent *"person"* in the image, the LLaVA-1.5 generates completely inaccurate responses, incorrectly describing a tennis-playing scene. However, LLaVA-1.5$_{ViHallu}$ successfully identifies the absence of people and accurately notes the specific detail of the ball on the racket. **These results demonstrate that ViHallu effectively reduces the model's reliance on language priors and significantly mitigates hallucination phenomena**. Further comparative examples illustrating model responses before and after fine-tuning with ViHallu can be found in the Appendix.

**Impact of the dataset size.** To evaluate the effect of fine-tuning dataset size on model performance, varying subsets are sampled from ViHallu-Instruction, ranging from 50 to 2,000 images with their corresponding instructions (averaging 7-8 instructions per image). The LVLMs are fine-tuned using these sampled datasets and evaluated on the POPE benchmark. As shown in Figure 13 (a), all LVLMs exhibit a general upward trend in performance that gradually plateaus as the amount of fine-tuning data increases. **This trend demonstrates that ViHallu effectively mitigates hallucination in LVLMs with increasing dataset size**.

**Consistency analysis with different subsets.** To assess the consistency of ViHallu-Instruction, four independent subsets are extracted from the dataset, each comprising 4,201 instructions paired with approximately 500 distinct images. Using these subsets for fine-tuning LLaVA-1.5, evaluations on the POPE benchmark reveal consistent performance improvements. As shown in Figure 13 (b), all fine-tuned models demonstrate substantial improvements over the baseline, with variations in the F1 score within 0.7 between subsets. This minimal variance suggests **the robustness of ViHallu-Instruction across different dataset samples**.

## 6 Conclusion

In this paper, we present ViHallu, a vision-centric hallucination mitigation framework that enhances visual-semantic alignment through visual variation image generation and visual instruction construction. ViHallu addresses a critical limitation in existing hallucination mitigation methods by introducing visual variation samples focuses on fine-grained visual differences. Extensive experiments across multiple benchmarks and LVLMs demonstrate that ViHallu effectively enhances models' fine-grained visual understanding while significantly reducing hallucination tendencies. Furthermore, ViHallu-Instruction dataset provides a valuable resource for future research in visual-semantic alignment and hallucination mitigation.

## Acknowledgments

This work was supported by Shanghai university.

(2023).

## A Overview

In this supplementary material, we present construction details of ViHallu-Instruction and additional examples.

## B ViHallu-Instruction

### B.1 Visual Variation Image Quality Evaluation

In this section, we present the quality assessment details for visual variation images. We employ VQAScore as our evaluation metric, which operates by converting image captions into questions. These questions, along with their corresponding images, are fed into an image-question encoder, where the answer decoder outputs a probability of "yes" as the quality score for visual variation images. VQAScore, built upon LVLMs, demonstrates robust Visual Question Answering (VQA) capabilities, making it particularly suitable for screening the quality of our generated visual variation images. During the construction of ViHallu-Instruction, we generate 7,209 visual variation images from 2,000 original images. Figure 8 illustrates the VQAScore distribution evaluated by LLaVA-1.5-13B. To ensure high-quality image selection, we establish a VQAScore threshold of 0.6, retaining images with scores greater than or equal to 0.6 while discarding those below this threshold. Due to consistently low-quality generations for some source images, their corresponding visual variation images are entirely excluded. Ultimately, the image component of ViHallu-Instruction comprises 5,051 visual variation images paired with 1,719 original images.

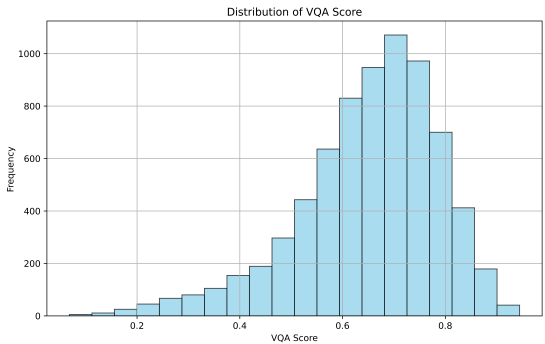

Figure 8: Distribution of VQAScore evaluated by LLaVA-1.5-13B for visual variation images. The threshold of 0.6 is used to filter high-quality images.

### B.2 Question Generation

This section details the specific steps for question generation based on detailed descriptions and object tags. We employ the DeepSeek-chat V2 model as a robust question generator, encompassing diverse question types. Specifically, we design a prompt template as illustrated in Figure 10, where contextual detailed descriptions and object tags are inserted into designated slots to generate corresponding questions. The generated questions span various types, effectively addressing different forms of hallucinations (including category, attribute, and relation hallucinations). These questions not only effectively reflect meaningful semantic information from the detailed descriptions but also challenge potential hallucinations within them. Additionally, we incorporate negative questions to enhance the diversity of the question set.

### B.3 QA Pairs Quality Assessment

To evaluate instruction quality, we employ a panel of LVLM experts (LLaVA-1.5, MiniCPM-V 2.6, and mPLUG-OWL3) to assess the alignment between QA pairs and corresponding images. The prompt template for this binary assessment ("Yes"/"No") is shown in Figure 11. QA pairs require endorsement from at least two expert models to be retained. The distribution of quality-assured and quality-filtered QA pairs from different baseline models is presented in Figure 9.

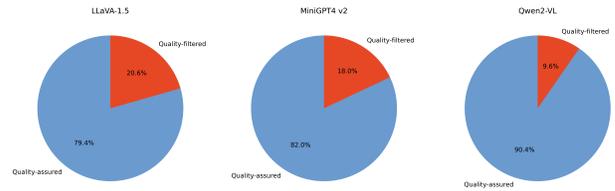

Figure 9: Pie chart represents the proportion of QA pairs that passed or failed expert validation from different baseline model.

## C Additional Examples

### C.1 Samples from ViHallu-Instruction

Figure 12 showcases sample pairs from ViHallu-Instruction, each consisting of an original image and its visual variation image along with their QA pairs. Despite their high visual similarity, these image pairs contain subtle yet distinguishing differences, resulting in distinct answers to identical questions. These samples demonstrate how ViHallu-Instruction effectively captures fine-grained visual variations through targeted question-answer pairs.

### C.2 Sample Model Outputs

We compare the outputs of base models ( LLaVA-1.5, MiniGPT-4 v2 and Qwen2-VL) and their ViHallu-instruction fine-tuned versions ( LLaVA-1.5$_{ViHallu}$, MiniGPT-4 v2$_{ViHallu}$ and Qwen2-VL$_{ViHallu}$) across various images. As shown in Figure 13 and Figure 14, the visualization results demonstrate that ViHallu-instruction effectively helps LVLMs mitigate hallucinated descriptions while enhancing the generation of reliable and detailed content.

**Symstem mesage**

You are a language assistant that helps to ask questions about a sentence and some entries.

---

**Prompt**

Given a caption and entries, you need to come up with seven questions based on the sentence and entries. These questions should help verify the factuality of the sentence.

The questions may concern the existence and basic attributes of the entities, such as color, action, etc., as well as the quantity and positional relationship of the entities, When inquiring about positional relationships, ensure the question is clear.

Keep the questions simple and no more than 20 words.

When outputting, only one question should be given per line. Avoid repeating questions. For each line, first output &, then output the question.

**Example:**

**Sentence:**

The image depicts a large herd of zebras grazing in a grassy field. The zebras are scattered throughout the field, with some closer to the foreground and others further in the background. The field is dotted with trees, providing a natural and serene setting for the zebras.

**Entries:**

zebra wheat

**Output:**

How many zebras are in the herd?

What color is the field?

Are there trees in the field?

Are the zebra closer to the foreground or the background?

...

Figure 10: Prompt template for generating diverse question types using detailed descriptions and object tags.

**Symstem mesage**

You are an assistant that helps to verify the factuality of the answer

---

**Prompt**

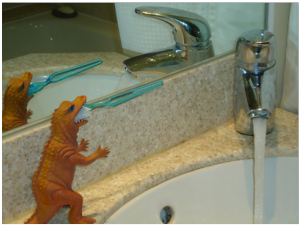

Verify whether the 'answer' in the following question-answer pairs is correct based on the image.

If the answer is correct, output 'yes';

If the answer is incorrect, output 'no'.

**Example:**

**Question:**

Is the toy alligator holding a toothbrush?

**Answer:**

Yes, the toy alligator is holding a toothbrush.

**Output:**

Yes

Figure 11: **Prompt template for LVLM expert evaluation of QA pairs. The template enables binary assessment of alignment between questions, answers, and image content.**

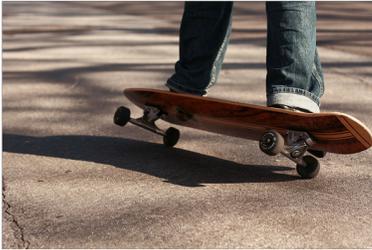
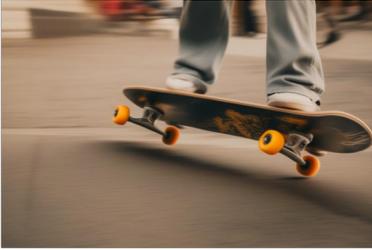

| Question | What is the surface the skateboarder is riding on? | What is the surface the skateboarder is riding on? |
| --- | --- | --- |
| Answer | The skateboarder is riding on a paved surface, likely asphalt. | The skateboarder is riding on a smooth, paved surface, likely a street or sidewalk. The surface appears even and flat, ideal for skateboarding. The motion blur suggests the skateboarder is moving quickly. The background is blurred, emphasizing the speed and movement. |

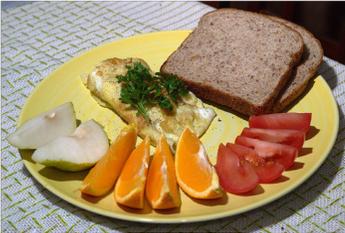
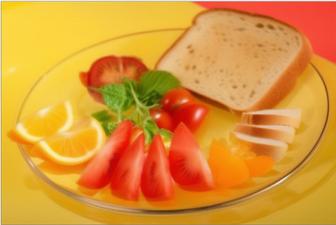

| Question | Is there a plate of food on the dining table? | Is there a plate of food on the dining table? |
| --- | --- | --- |
| Answer | Yes, there is a plate of food on the dining table. The plate contains a variety of items including slices of bread, an omelet garnished with parsley, slices of orange, and pieces of apple and tomato. The plate is yellow and is placed on a table with a green and white patterned tablecloth. | Yes, there is a plate of food on the dining table. The plate is set on a yellow surface, and the overall presentation is colorful and fresh, suggesting a healthy meal or snack. The combination of bread, fruit, and vegetables indicates a balanced and nutritious selection. |


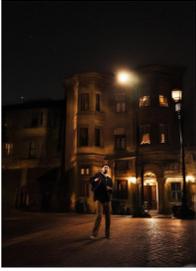

| Question | Are there other people in the image? | Are there other people in the image? |
| --- | --- | --- |
| Answer | Yes, there are other people walking along the sidewalk in the image. | No, there are no other people visible in the image. The photo shows a single person standing in front of a building at night. |

Figure 12: Sample pairs from ViHallu-Instruction, showing original and visual variation images with their corresponding QA pairs.

**Where is the girl positioned among the three people in the image**

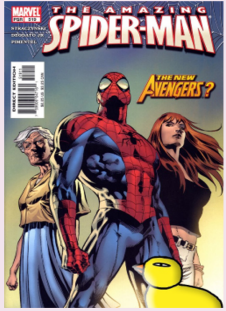

LLaVA-1.5: The girl is positioned between the two men in the image.

LLaVA-1.5*ViHallu*: In the image, the girl is positioned on the right side among the three people. She is standing next to the person in the middle, who is near a duck.

MiniGPT-4-V2: The girl is positioned between the two men in the image, one of which is on the left and the other is on the right.

MiniGPT-4-V2*ViHallu*: The girl is positioned on the right side of the image.

Qwen-VL: The girl is positioned to the right of the man and the woman in the image.

Qwen-VL*ViHallu*: The girl is positioned to the right of the elderly man and Spider-Man in the image. She is standing next to Spider-Man, with her body facing towards him. The elderly man is on the left side of the image, and he is standing behind Spider-Man.

Figure 13: Sample outputs from base models and their ViHallu fine-tuned counterparts

Which cat (left, right or middle) in the image opens its mouth?

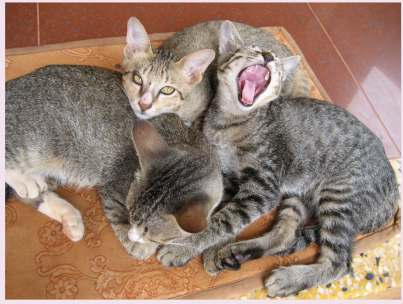

LLaVA-1.5: The middle cat in the image opens its mouth.

LLaVA-1.5*ViHallu*: The cat on the right opens its mouth.

MiniGPT-4-V2: middle

MiniGPT-4-V2*ViHallu*: The cat on the right side of the image opens its mouth.

Qwen-VL: The cat on the right side of the image opens its mouth.

Qwen-VL*ViHallu*: The cat on the right side of the image is the one that has its mouth open. It appears to be yawning or meowing. The other two cats have their mouths closed.

Figure 14: Sample outputs from base models and their ViHallu fine-tuned counterparts